\documentclass[conference,letterpaper]{IEEEtran}

\usepackage[utf8]{inputenc} 
\usepackage{graphicx}
\usepackage[T1]{fontenc}
\usepackage{url}
\usepackage{hyperref}
\usepackage{xcolor}
\usepackage{ifthen}
\usepackage{cite}
\usepackage{amssymb}
\usepackage{amsthm}
\usepackage{subfigure}
\usepackage{multicol}
\usepackage[cmex10]{amsmath} 

\DeclareMathOperator*{\argmin}{argmin}

\newtheorem{definition}{\bf{Definition}}

\newtheorem{theorem}{\bf{Theorem}}

\newtheorem{proposition}{\bf{Proposition}}
\newtheorem{remark}{\bf{Remark}}
\newtheorem*{proof2*}{\bf{Proof Sketch}}
\newtheorem{assumption}{\bf{Assumption}}

\begin{document}
\title{Online Transfer Learning: Negative Transfer and Effect of Prior Knowledge} 


\author{%

  \IEEEauthorblockN{Xuetong Wu$^1$, Jonathan H. Manton$^1$, Uwe Aickelin$^2$, Jingge Zhu$^{1}$}
  \IEEEauthorblockA{$^1$Department of Electrical and Electronic Engineering\\
                    $^2$Department of Computing and Information Systems\\
                    University of Melbourne\\ 
                    Parkville, Victoria, Australia\\
                    Email: xuetongw1@student.unimelb.edu, \{jmanton, uwe.aickelin, jingge.zhu\}.unimelb.edu.au}
}


\maketitle

\begin{abstract}
Transfer learning is a machine learning paradigm where the knowledge from one task is utilized to resolve the problem in a related task. On the one hand, it is conceivable that knowledge from one task could be useful for solving a related problem. On the other hand, it is also recognized that if not executed properly, transfer learning algorithms could in fact impair the learning performance instead of improving it - commonly known as \textit{negative transfer.} In this paper, we study the online transfer learning problems where the source samples are given in an off-line way while the target samples arrive sequentially. We define the expected regret of the online transfer learning problem, and provide upper bounds on the regret using information-theoretic quantities.  We also obtain exact expressions for the bounds when the sample size becomes large. Examples show that the derived bounds are accurate even for small sample sizes. Furthermore, the obtained bounds give valuable insight on the effect of prior knowledge for transfer learning in our formulation. In particular, we formally characterize the conditions under which negative transfer occurs.
\end{abstract}


\section{Introduction}
\begin{figure}[h!]
    \centering
    \includegraphics[width = 6cm]{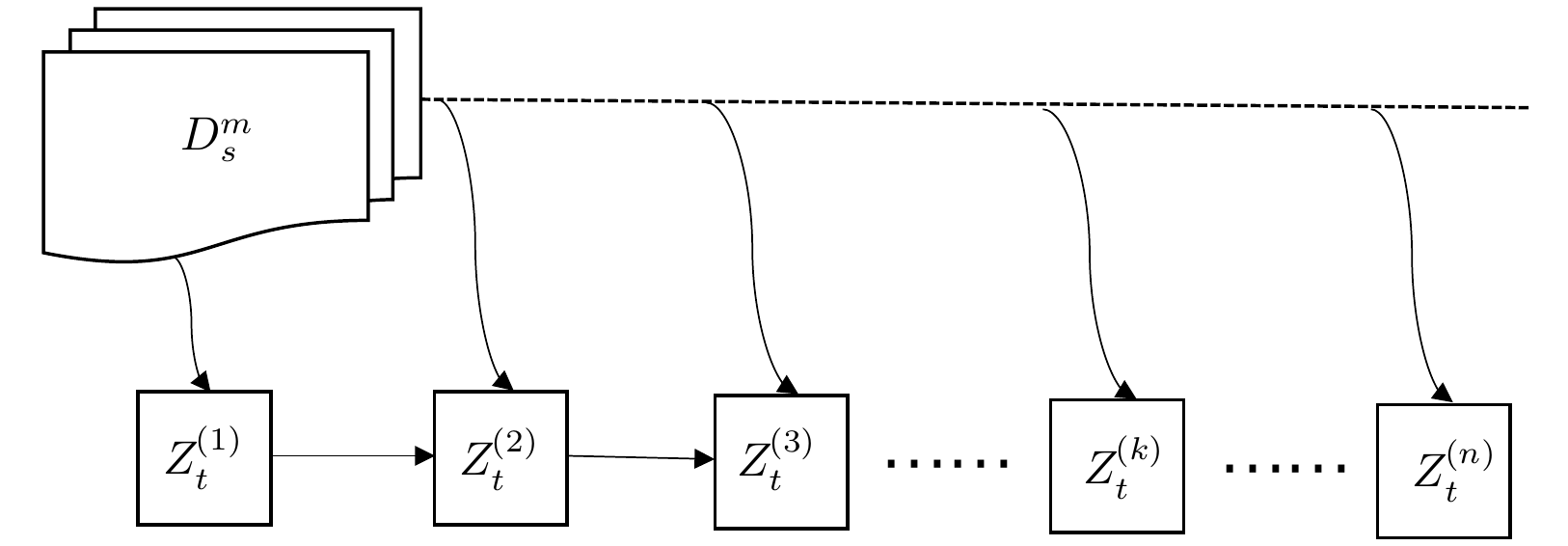}
    \caption{Online Transfer Learning Framework}
    \label{fig:otl}
\end{figure}

Transfer learning is a rising machine learning problem that leverages past knowledge in one or more \emph{source} tasks to resolve the problem (or improve the performance) in a related \emph{target} domain. The key problems are how to use the source intelligently to improve the performance in the target domain, and, how to characterize and avoid negative transfer. Currently, most existing transfer learning methods focus on offline settings where batch target data are available (see \cite{pan2009survey,weiss2016survey,zhuang2020comprehensive} and references therein). As such an assumption may not always hold in some real-time applications such as data transmission, we investigate the \emph{online transfer learning} that is firstly proposed by \cite{zhao2014online}. Unlike traditional online learning, the framework of OTL is illustrated in Figure~\ref{fig:otl}, where the decision is sequentially made with the aid of source data and  historical target data. 
This framework has been extended to many other problems such as multisource transfer \cite{wu2017online},\cite{kang2020online}, multi-task problem \cite{he2011graphbased} and iterative domain adaptation \cite{bhatt2016multi}. Yet most of these works do not focus on rigorous theoretical analysis. Moreover, they mostly consider specific learning tasks such as binary classifications with linear models, and the performance is evaluated using a very specific metric (e.g., the number of mistakes). Such a learning framework in general does not exploit the structures (or distributions) of the data or model parameters. Lastly, no prior work has formally studied the problem of negative transfer.

In this work, we propose a more general framework for the online transfer problem that is suitable for general setups from the information-theoretic view. The information-theoretic framework has been established and studied in many online learning and reinforcement learning problems \cite{lazaric2008transfer,lazaric2012transfer,zhan2015online,taylor2009transfer, merhav1998universal,russo2016information}. One advantage of this framework is that information-theoretic tools are quite useful in studying the asymptotic behaviors as well as deriving learning performance bounds for various statistical problems. This paper is inspired by the universal prediction framework \cite{merhav1998universal}. By universal we mean that no matter which distribution the data are drawn from, the predictor will always yield good performance with theoretical guarantees. Specifically, we formulate the online transfer learning problem under the assumption that the source and target data distributions are parameterized by some unknown parameters $\theta^*_s, \theta^*_t \in \Lambda$. Then we define the expected regret and further propose the "mixture" strategy for sequential target predictors. The asymptotic upper bounds are also derived for the expected regret. Practically, the bound can be also applied to the typical transfer learning regime where the abundant source data are available but the target data are lacking. To conclude, the contributions of this paper are listed as follows.
\begin{itemize}
    \item Consider the online transfer learning framework, at each time $k$, we propose the mixture strategy for the predictor $b_k$ with the prior knowledge over source and target parameters. Then the expected regret is characterized by the conditional mutual information (CMI).
    \item We give an asymptotic estimation of CMI for $\Lambda \subseteq \mathbb{R}^d$, where the bound is captured by the prior knowledge, the number of common parameters that $\theta^*_t$ and $\theta^*_s$ share, and the structure of the parametric family. The results can be easily extended to time-variant target domains.
    \item  Based upon the general asymptotic bounds, we show that the inappropriate prior will lead to the negative transfer. That is, using data from the source domain would hurt the performance on the target domain. To our best knowledge, this is the first theoretical study on the negative transfer perspective of online transfer learning.
    \item The logistic regression example is examined and simple experiments confirm the effects of the prior and show that the improper prior will lead to negative transfer.
\end{itemize}

\section{Problem formulation and main results}
Assume the source data $ D^m_s = (Z^{(1)}_s,\cdots,Z^{(m)}_s) \in \mathcal{Z}^{m}$ are given in batch while the target data are received sequentially as $Z^{(1)}_t, Z^{(2)}_t, \cdots, Z^{(k)}_t, \cdots $ where each sample takes value in $\mathcal{Z}$. Note that $\mathcal{Z}$ can be discrete or continuous. At each time instant $k$, after having seen $D^{k-1}_{t} = (Z^{(1)}_t, Z^{(2)}_t,\cdots, Z^{(k-1)}_t)$, we predict $Z^{(k)}_t$ using $D^m_s$ and $D^{k-1}_t$ with the predictor $b_k : \mathcal{Z}^{m} \times \mathcal{Z}^{k-1} \rightarrow \hat{\mathcal{Z}}$. Note that $\hat{\mathcal{Z}}$ could be different from $\mathcal{Z}$ in general, e.g., $\hat{\mathcal{Z}}$ is a quantized version of $\mathcal{Z}$. We define the loss function $\ell: \hat{\mathcal{Z}} \times \mathcal{Z} \rightarrow \mathbb{R}^{+}$ that evaluates the prediction performance. In this paper, we use the convention that capital letters denote the random variables and small letters as their realizations. We further make the following assumptions.
\begin{assumption}[Parametric Distributions]\label{asp:para-dist}
    We assume that source and target data are generated independently in an i.i.d. fashion. More precisely, the joint distribution of the data sequence pairs $P_{\theta^*_s,\theta^*_t}({D_t^n, D_s^m})$ can be written as 
    \begin{small}
    \begin{equation}
    P_{\theta^*_s,\theta^*_t}(D_t^n, D_s^m)= \prod_{i=1}^n P_{\theta_t^*}(Z_t^{(i)}) \prod_{j=1}^m P_{\theta_s^*}(Z_s^{(j)}),
    \end{equation}
    \end{small}
    where $P_{\theta_t^*}$ and $P_{\theta_s^*}$ are in a parametrized family of distributions $\mathcal{P} = \{P_{\theta}\}_{\theta \in \Lambda}$. Here $\Lambda \subseteq \mathbb{R}^{d}$ is some measurable space and $\theta^*_t$ and $\theta^*_s$ are points in the interior of $\Lambda$.
\end{assumption}
After observing $n$ target samples, we want to minimise the corresponding expected regret defined as,
\begin{small}
\begin{align}
    \mathcal{R}(n) := \mathbb{E}_{\theta^*_s,\theta^*_t} \left[\sum_{k=1}^{n} \ell\left(b_{k}, Z^{(k)}_t\right) - \sum_{k=1}^{n} \ell(b_k^*, Z^{(k)}_t) \right],
\end{align}
\end{small}
where $b_1$ is learned from  $D^m_s$ and $b_k (k>1)$ is the decision we made based on both  $D^m_s$ and  $D^{k-1}_t$ but without the knowledge of $\theta^*_s$ and $\theta^*_t$ . The predictor $b_k^*$ is the optimal decision at each time $k$ that can depend on true target distributions $P_{\theta^*_t}$. If not otherwise specified, the notation $\mathbb{E}_{\theta_s,\theta_t}[\cdot]$ (similar to $\mathbb{E}_{\theta_t}[\cdot]$ and $\mathbb{E}_{\theta_s}[\cdot]$) means the expectation is taken over all source and target samples that are drawn from ${P}_{\theta_s}$ and ${P}_{\theta_t}$. 
\subsection{Expected Regret Bounds}
\noindent Given the above problem formulation, we begin by considering the \emph{logarithm loss} defined as follows.
\begin{definition}[Logarithm Loss] Let the predictor $b_k$ be a probability distribution over the sample $z^{(k)}_t$ at time $k$, the logarithm loss is defined as 
\begin{small}
\begin{equation}
    \ell(b_k,z^{(k)}_t) = - \log b_k(z^{(k)}_t).
\end{equation}
\end{small}
\end{definition}
At each time $k$, we may view the predictor as a conditional probability distribution $b_k(z^{(k)}_t) = Q(z^{(k)}_t|D^{k-1}_t, D^{m}_s) $, and the optimal predictor $b^*_k$ is naturally given by the true target distribution over $z^{(k)}_t$ as $b^*_k(z^{(k)}_t) = P_{\theta^*_t}(z^{(k)}_t)$. Then the expected regret till time $n$ can be written explicitly as
\begin{small}
\begin{align}
   \mathcal{R}_{\textup{log}}(n) &= \mathbb{E}_{\theta^*_s,\theta^*_{t}}\left[  \log \frac{1}{Q(D^n_t| D^{m}_s)}-  \log \frac{1}{P_{\theta^*_t}(D^n_t)} \right].  \label{eq:logloss}
\end{align}
\end{small}
The effect of source data is reflected in the conditional distribution $Q(D^n_t| D^{m}_s)$. Concerning the choice of $Q$, we first define $\Theta_s$ and $\Theta_t$ as random variables over $\Lambda$.  Since $P_{\theta^*_s}$ and $P_{\theta^*_t}$ are unknown, we assign a probability distribution $\omega$ over $\Theta_s$ and $\Theta_t$ w.r.t. the Lebesgue measure to represent our prior knowledge and update the posterior with the incoming data to approximate the underlying distributions, which is known as the \emph{mixture strategy}\cite{merhav1998universal,xie2000asymptotic}. In particular, we choose the predictor $Q(D^n_t|D^m_s)$ as
\begin{small}
\begin{align}
  Q(D^n_t|D^m_s) &= \frac{\int P_{\theta_t,\theta_s}(D^n_t, D^m_s)\omega(\theta_t, \theta_s)d \theta_t d\theta_s}{\int P_{\theta_s}(D^m_s)\omega(\theta_s)d\theta_s} \nonumber \\
  &= \int P_{\theta_t}(D^n_t)\omega(\theta_t|\theta_s)  d\theta_t Q(\theta_s|D^m_s) d\theta_s, \label{eq:mixture}
\end{align}
\end{small}
where $\omega(\theta_s)$ is the marginal of $\omega(\theta_s, \theta_t)$. From Eq~(\ref{eq:mixture}), the mixture strategy quantitatively explains how transfer learning is implemented via the posterior updates of $\theta_t$ from a Bayesian perspective. Intuitively speaking, the posterior $Q(\theta_s|D^m_s)$ firstly gives an estimate of $\theta^*_s$ from the source data, then the conditional prior $\omega(\theta_t|\theta_s)$ reflects our belief upon $\theta^*_t$ given $\theta_s$ estimated from source data. With the choice of $Q(D^n_t|D^m_s)$, the expected regret can be explicitly characterized in the following theorem.
\begin{theorem} [Regret with Log-loss]\label{thm:cmi}
With the mixture strategy $Q(D^n_t|D^m_s)$, the expected regret in Eq~(\ref{eq:logloss}) can be written as
\begin{small}
\begin{align}
   \mathcal{R}_{\textup{log}}(n) = \mathbb{E}_{\theta^*_s,\theta^*_t}\left[ \log\frac{P_{\theta^*_t}(D^n_t)}{Q(D^n_t
   |D^m_s)} \right] = I(D^n_t;\Theta_t = \theta^*_t, \Theta_s = \theta^*_s|D^{m}_s), \label{eq:cmi} 
\end{align}
\end{small}
where $I(D^n_t;\Theta_t = \theta^*_t, \Theta_s = \theta^*_s|D^{m}_s)$ denotes the conditional mutual information $I(D^n_t;\Theta_t, \Theta_s | D^{m}_s)$ evaluated at $\Theta_t = \theta^*_t, \Theta_s = \theta^*_s$.
\end{theorem}
\noindent All proofs in this paper can be found in \cite{Proofs_2021}. In many cases, we need to consider the task-specific loss such as squared loss, 0-1 loss and hinge, etc. For other general bounded loss function $\ell$, we define the predictor at time $k$ to be,
\begin{small}
\begin{equation}
    b_k = \argmin_{b} \mathbb{E}_{Q(D^k_t,D^{m}_s)}\left[ \ell(b,z^{(k)}_t)|D^m_s,D^{k-1}_t \right], \label{eq:general_bk}
\end{equation}
\end{small}
with the choice of the mixture strategy $Q(D^k_t,D^{m}_s) = \int P_{\theta_t,\theta_s}(D^k_t, D^m_s)\omega(\theta_t, \theta_s)d \theta_t d\theta_s$ for some prior $\omega$. The optimal predictor is naturally given by,
\begin{small}
\begin{equation}
b_k^* = \argmin_{b} \mathbb{E}_{P_{\theta^*_t}(D^k_t)}\left[ \ell(b,z^{(k)}_t)|D^{k-1}_t \right]. \label{eq:general_bkstar}
\end{equation}
\end{small}
As a consequence, we arrive at the following theorem.
\begin{theorem}[Bounds on General Loss]\label{thm:general-loss}
Assume the loss function satisfies $|\ell(b,z) - \ell(b^*,z)| \leq M$ for any observation $z$ and the predictors $b,b^*$. Then the true expected regret induced by $b_k$ and $b^*_k$ in Eq~(\ref{eq:general_bk}) and~(\ref{eq:general_bkstar}) can be bounded as,
 \begin{small}
\begin{equation}
    \mathcal{R}(n) \leq M\sqrt{2n I(D^n_t;\Theta_t = \theta^*_t, \Theta_s = \theta^*_s|D^{m}_s)}. \label{ineq:bound}
\end{equation}
\end{small}
\end{theorem}
The above theorem implies that if the loss function is bounded, with a certain prior $\omega$, the regret induced by the mixture strategy is also captured by CMI evaluated at $\Theta = \theta^*_t$ and  $\Theta_s = \theta^*_s$. However, the bound in its current form is less informative as it does not show what is the effect of the prior $\omega$ and  sample size $m$ and $n$.

\subsection{Asymptotic Analysis of CMI}
To further investigate the effect of prior, we give an asymptotic analysis of CMI. First we make the regular assumptions on parametric conditions \cite{clarke1999asymptotic, clarke1990information} and define the proper prior.
\begin{assumption}[Parametric Condition] \label{asp:para-trans}
Assume the source and target distributions $P_{\theta^*_s}(Z_s)$ and $P_{\theta^*_t}(Z_t)$ are twice continuously differentiable at $\theta^*_s$ and $\theta^*_t$ for almost every $Z_s$ and $Z_t$. For any $\theta_t, \theta_s \in \Lambda $, there exist $\delta_s, \delta_t > 0$ satisfying, 
\begin{footnotesize}
\begin{align}
    \mathbb{E}_{\theta_t} \left[ \sup _{\left\|\theta_t-\theta^*_t\right\| \leq \delta}\left|\frac{\partial}{\partial \theta_{t,i}} \log P_{\theta_t}\left(Z_t \right)\right| \right] &< \infty \\
    \mathbb{E}_{\theta_s} \left[ \sup _{\left\|\theta_s-\theta^*_s\right\| \leq \delta}\left|\frac{\partial}{\partial \theta_{s,i}} \log P_{\theta_s}\left(Z_s \right)\right| \right] &< \infty  
\end{align}
\end{footnotesize}
for $i = 1,\cdots, d$. In addition, we assume,
\begin{footnotesize}
\begin{align}
\mathbb{E}_{\theta_t} \left[ \sup _{\left \|\theta_t-\theta^*_t\right\| \leq \delta}\left|\frac{\partial^{2}}{\partial \theta_{i} \partial \theta_{j}} \log P_{\theta_t}\left(Z_t \right)\right|^{2}\right] <\infty \\
\mathbb{E}_{\theta_s} \left[ \sup _{\left\|\theta_s-\theta^*_s\right\| \leq \delta}\left|\frac{\partial^{2}}{\partial \theta_{i} \partial \theta_{j}} \log P_{\theta_s}\left(Z_s \right)\right|^{2}\right] <\infty 
\end{align}
\end{footnotesize}
for any $i,j = 1,\cdots, d$. 
\end{assumption}
\begin{definition}[Proper Prior] \label{def:proper-prior}
Given a prior $\omega(\Theta_s, \Theta_t)$, we say,
\begin{itemize}
    \item the induced marginal density $\omega(\Theta_s)$ is proper if it is continuous and positive over the whole support $\Lambda \subseteq \mathbb{R}^d$.
    \item the conditional density $\omega(\Theta_t|\Theta_s)$ is proper if there exist some $\delta_s > 0$ and $\delta_t > 0$ such that $\omega(\theta_t|\theta_s) > 0$
    for any $\theta_s$ and $\theta_t$ satisfying $\|\theta_s - \theta^*_s\| \leq \delta_s$ and $\|\theta_t - \theta^*_t\| \leq \delta_t$. 
    \item $\omega(\Theta_s, \Theta_t)$ is proper if $\omega(\Theta_s)$ and $\omega(\Theta_t|\Theta_s)$ are proper.
\end{itemize}
We also define the proper prior without the source as having the continuous density $\hat{\omega}(\theta_t) > 0$ over the whole support $\Lambda$.
\end{definition}
If the distributions in parametric family $\mathcal{P}_{\theta}$ satisfy the Assumption~\ref{asp:para-trans} (e.g., the exponential families in \cite{takeuchi1998asymptotically}), with the proper prior, we ensure that the posterior distribution of $\Theta_t$ and $\Theta_s$ given $D^{n}_t$ and $D^{m}_s$ asymptotically concentrates on neighborhoods of $\theta^*_{t}$ and $\theta^*_s$, respectively. With definitions in place, for the case where both $\Theta_s$ and $\Theta_t$ are scalars, we give the asymptotic estimation for CMI as follows.
\begin{theorem}[Asymptotic Estimation of CMI]\label{thm:consistency-scalar}
 Under Assumptions~\ref{asp:para-dist} and \ref{asp:para-trans}, for $\Lambda = \mathbb{R}$ and $\theta^*_s \neq \theta^*_t$,   as $n, m \rightarrow \infty$, the mixture strategy with proper prior $\omega(\Theta_s,\Theta_t)$ yields,
 \begin{footnotesize}
\begin{align}
    I(D^n_t;  \Theta_t  = \theta^*_t,   \Theta_s = \theta^*_s|  D^{m}_s) - \frac{1}{2}\log \frac{n}{2\pi e}  \rightarrow    \frac{1}{2}\log I_t(\theta^*_t)  + \log\frac{1}{\omega(\theta^*_t|\theta^*_s)}, \label{eq:withsource}
\end{align}
 \end{footnotesize}
where we define the Fisher information $\mathbb{E}_{\Theta_t} \left[ - \nabla^2_{\Theta_t} \log P_{\Theta_t}(Z_t)\right]$ evaluated at $\Theta_t = \theta^*_t$ as $I_t(\theta^*_t)$.
\end{theorem}
\begin{remark} \label{remark:consist}
Compared to the result without the source data when target sample is abundant\cite{clarke1990information},
\begin{small}
\begin{align}
    I(D^n_t;\Theta_t = \theta^*_t) - \frac{1}{2}\log \frac{n}{2\pi e} \rightarrow    \frac{1}{2}\log I_t(\theta^*_t)  + \log\frac{1}{\hat{\omega}(\theta^*_t)} \label{eq:withoutsource}
\end{align}
\end{small}
for some prior $\hat{\omega}(\Theta_t)$, the difference between Eq~(\ref{eq:withsource}) and (\ref{eq:withoutsource}) is $\frac{\hat{\omega}(\theta^*_t)}{\omega(\theta^*_t|\theta^*_s)}$. It says that if the distribution $\omega$ can be chosen such that $\frac{\hat{\omega}(\theta^*_t)}{\omega(\theta^*_t|\theta^*_s)} < 1$, the source data will help to reduce the regret. However, it should be noted that $\omega$ is chosen without knowing the exact value of $\theta^*_t$ and $\theta^*_s$ so it is not immediately clear if this is always possible. We will show later that if the conditional prior $\omega(\theta_t|\theta_s)$ is proper, it is always possible to find a distribution such that $\frac{\hat{\omega}(\theta^*_t)}{\omega(\theta^*_t|\theta^*_s)} < 1 $. On the contrary, if the prior information between the source and target is incorrect, we may always end up with $\frac{\hat{\omega}(\theta^*_t)}{\omega(\theta^*_t|\theta^*_s)} > 1$, which is one way to interpret {\bf negative transfer}.
\end{remark}
\begin{remark}
Notice that the source samples change the constant from $\log\frac{1}{\hat{\omega}(\theta_t)}$ to $\log\frac{1}{\omega(\theta_t|\theta_s)}$, which are independent from $n$. Hence the effect of the source samples vanishes asymptotically as $n$ goes to infinity. However, the asymptotic analysis is still useful for two reasons. Firstly, we will show later that when both $n$ and $m$ approach infinity, the sample complexity of the regret (i.e., how regret scales in terms of $m$ and $n$) can change, depending on how fast $m$ and $n$ grow relative to each other. Secondly, our numerical results show that the asymptotic bound is in fact very accurate even for relatively small $m$ and $n$.
\end{remark}

Theorem~\ref{thm:consistency-scalar} holds when the distributions are parametrized by scalars. We extend to a more typical transfer learning scenario where $\Theta_t, \Theta_s \in \mathbb{R}^d$ with $d > 1$ share  some common parameters $\Theta_c \in \mathbb{R}^j$ for $0 \leq j \leq d$. To illustrate, we can write the parameters in the following way.
\begin{small}
\begin{align*}
    \Theta_s &= (\Theta_{c,1}, \Theta_{c,2},\cdots, \Theta_{c,j}, \quad \Theta_{s,1}, \cdots, \Theta_{s,d-j}  ) =  (\Theta_c, \Theta_{sr}) \\
    \Theta_t &= (\underbrace{\Theta_{c,1}, \Theta_{c,2},\cdots, \Theta_{c,j}}_{\text{common parameters}}, \quad \underbrace{\Theta_{t,1}, \cdots, \Theta_{t,d-j})}_{\text{task-specific parameters}} =  (\Theta_c, \Theta_{tr})
\end{align*}
\end{small}
where $\Theta_c \in \mathbb{R}^{j}$ denotes the common parameter vector and $\Theta_{sr} , \Theta_{tr} \in \mathbb{R}^{d-j}$ are task-specific parameter vectors. Then we reach the following theorem that gives the asymptotic normality of the conditional mutual information with $d > 1$.
\begin{theorem}[Asymptotic Estimation for General Parametrization]\label{theorem:gene-para}
 Under Assumptions~\ref{asp:para-dist} and \ref{asp:para-trans}, with $\Theta_s,\Theta_t \in \mathbb{R}^d$ defined above and as $n, m \rightarrow \infty$, the mixture strategy with proper prior $\omega(\Theta_s,\Theta_t)$ yields,
 \begin{small}
\begin{align}
&I(D^n_t;\Theta_t = \theta^*_t, \Theta_s = \theta^*_s|D^{m}_s)  - \frac{1}{2} \log \operatorname{det}(\mathbf{I}_{j\times j} + \frac{n}{m} \Delta_t \Delta^{-1}_s)  \nonumber \\
&  - \frac{1}{2} \log \operatorname{det}(n I_t(\theta^{*}_{tr}))  \rightarrow  \frac{d-j}{2}\log \frac{1}{2\pi e} + \log \frac{1}{\omega(\theta^*_t|\theta^*_s)},
\end{align}
 \end{small}
where $\Delta_s = I_{cs}(\theta^*_c) - I_{cs}(\theta^*_c,\theta^{*}_{sr}) I^{-1}_s(\theta^{*}_{sr}) I^{T}_{cs}(\theta^*_c,\theta^{*}_{sr})$ and $\Delta_t = I_{ct}(\theta^*_c) - I_{ct}(\theta^*_c,\theta^{*}_{tr}) I^{-1}_t(\theta^{*}_{tr}) I^{T}_{ct}(\theta^*_c,\theta^{*}_{tr}) $, $\mathbf{I}_{j\times j}$ denotes the identity matrix with size $j$ and $\boldsymbol \theta^* = (\theta^*_c, \theta^{*}_{sr}, \theta^{*}_{tr})$ denotes the true parameters. With a little abuse of notation, we define the fisher information matrix as 
\begin{footnotesize}
\begin{align*}
I_{cs}(\theta^*_c) &= - \mathbb{E}_{\theta^*_s}\left[ \nabla^2_{\Theta_c}  \log P(Z_s |\Theta_c, \theta^*_{sr}) \right]\Big|_{\Theta_c = \theta^*_c} \in \mathbb{R}^{j\times j} \\
I_{ct}(\theta^*_c) &= - \mathbb{E}_{\theta^*_t}\left[ \nabla^2_{\Theta_c}  \log P(Z_t |\Theta_c, \theta^*_{tr}) \right]\Big|_{\Theta_c = \theta^*_c} \in \mathbb{R}^{j\times j}\\
I_{s}(\theta^*_{sr}) &= - \mathbb{E}_{\theta^*_s}\left[ \nabla^2_{\Theta_{sr}}  \log P(Z_s |\theta^*_c, \Theta_{sr})\right]\Big|_{\Theta_{sr} = \theta^*_{sr}} \in \mathbb{R}^{(d-j) \times (d-j)} \\
I_{t}(\theta^*_{tr}) &= - \mathbb{E}_{\theta^*_t}\left[ \nabla^2_{\Theta_{tr}}  \log P(Z_t |\theta^*_c, \Theta_{tr})\right]\Big|_{\Theta_{tr} = \theta^*_{tr}} \in \mathbb{R}^{(d-j) \times (d-j)}\\
I_{cs}(\theta^*_c, \theta^*_{sr}) &= -\mathbb{E}_{\theta^*_s}\left[\frac{\partial\log P(Z_s |\theta^*_c, \theta^*_{sr})}{  \partial \Theta_{c,i} \partial \Theta_{sr,k}} \right]_{ \begin{matrix}
i = 1,\cdots, j,\\
 k = 1,\cdots, d-j
 \end{matrix}} \in \mathbb{R}^{j\times (d-j)} \\
I_{ct}(\theta^*_c, \theta^*_{tr}) &= - \mathbb{E}_{\theta^*_t}\left[\frac{\partial\log P(Z_t |\theta^*_c, \theta^*_{tr})}{\partial \Theta_{c,i} \partial \Theta_{tr,k}} \right]_{ \begin{matrix}
i = 1,\cdots, j,\\
 k = 1,\cdots, d-j
 \end{matrix}} \in \mathbb{R}^{j \times (d-j)}
\end{align*}
\end{footnotesize}
\end{theorem}
\begin{remark}\label{remark:knowledge_transfer}
In the above expression, we can intuitively interpret the term $\frac{1}{2} \log \operatorname{det}(\mathbf{I}_{j\times j} + \frac{n}{m} \Delta_t \Delta^{-1}_s)$ as the "learning cost" of $\theta_c$, which is captured by the ratio $\frac{n}{m}$. If $m$ is
\begin{itemize}
    \item sublinear in $n$, the rate is $O(\log(nj))$ and source samples do not improve the learning performance 
    \item linear in $n$, the cost reduces to $O(1)$. 
    \item superlinear in $n$, the rate is $o(1)$, and abundant source samples indeed improve the performance and the cost vanishes in this case.
\end{itemize}
While the learning cost of $\theta_{tr}$ is relied on $\frac{1}{2} \log \operatorname{det}(n I_t(\theta^{*}_{tr}))$ and the prior $\omega(\theta^*_t|\theta^*_s)$, whereas the prior knowledge can only change the constant but does not change the rate.
\end{remark}

\begin{remark}
As a special case, if there is no common parameters ($j = 0$), then as both $m$ and $n$ are sufficiently large, 
\begin{small}
\begin{align}
    I(D^n_t; \Theta_t = \theta^*_t, & \Theta_s = \theta^*_s|D^{m}_s)   - \frac{d}{2}\log \frac{n}{2\pi e}\nonumber \\
    &\rightarrow   \frac{1}{2}\log  \operatorname{det}(I_t(\theta^*_{tr}))  +  \log \frac{1}{\omega(\theta^*_{t}|\theta^*_{s})}. \nonumber
\end{align}
\end{small}
Let $d = 1$, we can recover the results in Theorem~\ref{thm:consistency-scalar} and the knowledge transfer is only reflected on the prior knowledge $\omega(\theta^*_t|\theta^*_s)$. If the number of the common parameters is $d$ ($j = d$), that is, the source and target distributions are characterized by the same parameters, which yields the asymptotic estimation as,
\begin{small}
\begin{align}
   I(D^n_t; \Theta_t = \theta^*_t, \Theta_s = \theta^*_s|D^{m}_s) & - \frac{1}{2}\log \operatorname{det}(\mathbf{I}_{d \times d} + \frac{n}{m}I_{ct}(\theta^*_c)I^{-1}_{cs}(\theta^*_c)) \nonumber\\
    & \rightarrow  \log\frac{1}{\omega(\theta^*_t|\theta^*_s)} \nonumber.
\end{align}
\end{small}
Under this case, the regret depends on the ratio $\frac{n}{m}$ and prior $\omega(\theta^*_t|\theta^*_s)$ as discussed in Remark~\ref{remark:knowledge_transfer}.
\end{remark}

\subsection{Time-variant Target Domains}
 In the above problem, we assume that the target parameter $\theta_t^*$ stays fixed for all time. However, in some applications, the target distribution may change over time, and this motivates us to consider the time-variant transfer learning scenarios. Let the time evolving target data be parametrized by $\theta^*_{t,l}$ where at each index $l \in \mathbb{N}^+$, we will receive $n_l$ target samples $Z^{(i)}_{t,l}$ drawn from the distribution $P_{\theta^*_{t,l}}$. It is common to assume that $\theta^*_{t,l}$ only depends on the previous parameter $\theta^*_{t,l-1}$. At index $k$,  we are interested in minimising the expected regret
 \begin{small}
\begin{align}
    \mathcal{R}(k) = \sum_{l=1}^{k}\mathbb{E}_{\theta^*_s,\theta^*_{t,l},\theta^*_{t,l-1}} \left[\sum_{i=1}^{n_l} \ell\left(b_{i}, Z^{(i)}_{t,l}\right) - \sum_{i=1}^{n_l} \ell(b_i^*, Z^{(i)}_{t,l}) \right]. \label{time-variant:result}
\end{align} 
 \end{small}
Here $b_i$ is chosen to be the mixture strategy over $\theta^*_s,\theta^*_{t,l}$, and $\theta^*_{t,l-1}$. Combining Theorem~\ref{thm:general-loss} and~\ref{theorem:gene-para}, one can easily reach the asymptotic estimation of the expected regret.
\begin{theorem}[Time-variant Target Regret Bounds] Given the time-variant target domain described above, suppose that conditions in Therorem~\ref{thm:general-loss} and Assumptions~\ref{asp:para-dist} and \ref{asp:para-trans} hold for each $\theta^*_{t,k}$ and $\theta^*_s$. For $l = 1,2,\cdots,k$, we further assume that source parameters will share $j$ parameters with every $\theta^*_{t,l}$, and $\theta^*_{t,l}$, $\theta^*_{t,l-1}$ have $c_l$ common parameters. As $n_l, m \rightarrow \infty$, the mixture strategy with proper prior $\omega(\theta_s,\theta_{t,l}, \theta_{t,l-1})$ yields,
\begin{small}
\begin{align*}
    &\mathcal{R}(k) \leq M \Bigg(k \sum_{l=1}^{k} n_l \Big( \log\operatorname{det}\left(\mathbf{I}_{j \times j} + \frac{n_l}{m+n_{l-1}}\Delta_{ct}\Delta^{-1}_{cst} \right) \\
    &+ \log \operatorname{det}(\mathbf{I}_{c_l\times c_l} + \frac{n_l}{n_{l-1}} \Delta_{t} \Delta^{-1}_{t-1}) + \log\operatorname{det}(n_lI_{t,l}(\theta^*_{tr,l})) \\
    &+(d - j - c_l)\log\frac{1}{2\pi e} + \frac{2}{\omega(\theta^*_{t,l}|\theta^*_{t,l-1},\theta^*_s)}\Big) \Bigg)^{\frac{1}{2}}.
\end{align*}
\end{small}
\end{theorem}
In this case, the prior knowledge $\omega(\theta^*_{t,l}|\theta^*_{t,l-1},\theta^*_s)$ and the common parameters among which determine the prediction performance. Due to the space limit, we omit some analogous definitions and settings here, but readers can refer to the supplementary proof\cite{Proofs_2021} for more details and insights.
\subsection{Improper Prior and Negative Transfer}
As previously discussed, $\omega(\Theta_s,\Theta_t)$ should be chosen properly so that the posterior updating will asymptotically converge to the true parameter $\theta^*_s$ and $\theta^*_t$. However, if the prior distribution (particularly $\omega(\Theta_t|\Theta_s)$) is imposed improperly, the extra source data do not necessarily mean that our prediction for target data can always be improved. Roughly speaking, if our prior knowledge on $\theta_s^*$ and $\theta_t^*$ is incorrect, under our scheme, this would translate to an improper prior distribution for the mixture strategy. We will show that with an improper prior, the extra source data will in fact cause a higher regret (i. e. worse performance) compared to the case without source data.  
\begin{proposition}[Negative Transfer]\label{claim:negative}
Let $\mathcal{R}_{\omega(\Theta_s,\Theta_t)}(n,m)$ denote the regret induced by the mixture strategy  $Q(D^n_t|D^m_s)$ with the prior $\omega(\Theta_s,\Theta_t)$ and $\mathcal{R}_{\hat{\omega}(\Theta_t)}(n)$ denote the regret induced by $\hat{Q}(D^n_t)$ with the prior $\hat{\omega}(\Theta_t)$. If $\omega(\Theta_t|\Theta_s)$ is improper\footnote{We say $\omega(\Theta_t|\Theta_s)$ is improper if it does not satisfy conditions in Def~\ref{def:proper-prior}.}, then for any proper $\hat{\omega}(\Theta_t)$, the following inequality holds when both $n$ and $m$ are sufficiently large,
\begin{align}
    \mathcal{R}_{\omega(\Theta_s,\Theta_t)}(n,m) > \mathcal{R}_{\hat{\omega}(\Theta_t)}(n).
\end{align}
\end{proposition}
\begin{proof2*} 
\begin{itemize}
    \item By subtraction, we need to prove that,
    \begin{align*}
        \mathbb{E}_{\theta^*_t,\theta^*_s}\left[\log \frac{Q(D^m_s)\hat{Q}(D^n_t)}{Q(D^n_t, D^{m}_s)} \right] > 0,
    \end{align*}        
    \item Let us examine the logarithm term in the expectation as,
    \begin{small}
    \begin{align*}
    & \log \frac{Q(D^m_s)\hat{Q}(D^n_t)}{Q(D^n_t, D^{m}_s)}  =  \log \frac{1}{\int \int \hat{Q}(\theta_t|D^n_t)\frac{\omega(\theta_t|\theta_s)}{\hat{\omega}(\theta_t)}d \theta_t Q(\theta_s|D^m_s)d\theta_s} .
    \end{align*}
    \end{small}
    \item It can be found that the difference is characterized by the ratio  $\frac{\omega(\theta_t|\theta_s)}{\hat{\omega}(\theta_t)}$ and improper $\omega$ leads to zero mass near $\theta^*_t$ compared to proper $\hat{\omega}$, thus a higher regret.
\end{itemize}
\end{proof2*}
For example, let $\mathcal{Z} = \{0,1\}$ and assume $\Theta_s$ and $\Theta_t$ are the probabilities that the source and target samples take value in $1$. Also assume that our (incorrect) prior knowledge on the parameters is that $|\theta_s-\theta_t| \leq 0.1$ given any $\theta_s\in \Lambda$. Suppose the true underlying parameters are $\theta^*_t = 0.6$ and $\theta^*_s = 0.8$. In this case, 
even if knowing $\theta^*_s$ precisely, one can never end up with the correct estimation for $\theta^*_t$ even with abundant target samples if the prior $\omega(\theta_t|\theta_s)$ is improper. As a consequence, the regret becomes higher compared to the case without knowing such prior. For those who are interested in detailed analysis,  we refer to \cite{Proofs_2021} for more theoretical and experimental results. 

In contrast, if $\omega(\Theta_s,\Theta_t)$ is chosen properly, we can always find a prior such that the knowledge transfer from source data encourages lower regret, namely, the positive transfer.
\begin{proposition}[Positive Transfer]\label{claim:positive}
For any proper $\hat{\omega}(\Theta_t)$, there always exists a proper prior $\omega(\Theta_s,\Theta_t)$ that leads to the following inequality when both $n$ and $m$ are sufficiently large,
\begin{align}
    \mathcal{R}_{\omega(\Theta_s,\Theta_t)}(n,m) < \mathcal{R}_{\hat{\omega}(\Theta_t)}(n).
\end{align}
\end{proposition}
In our claim, we can always find a proper prior $\omega(\Theta_s,\Theta_t)$ whose marginal $\omega(\Theta_t|\Theta_s)$ encourages a tighter support over $\Theta_t$. In other words, making use of source data appropriately can narrow down the uncertainty range over $\Theta_t$. It then follows that such prior assigns more concentrated mass around $\theta^*_t$, which reduces the expected regret.

\section{Examples}\label{sec:examples}
 Consider a logistic regression problem in a 2-dimensional space. For the given parameter $\theta \in [0,1]^2$ and $Z_i = (X_i,Y_i) \in \mathbb{R}^{2} \times \{0,1\}$, each label $Y_i \in \{0, 1 \}$, is generated from a Bernoulli distribution with probability $p(Y_i = 1) = \frac{1}{1+e^{-\theta^TX_i}}$. Suppose that the source and target input features $X^{(k)}_s$ and $X^{(k)}_t$ are drawn from the same normal distribution $\mathcal{N}(\begin{bmatrix}
5  \\
-5
\end{bmatrix} ,\begin{bmatrix}
1 & 0\\
0 & 1
\end{bmatrix})$. The loss function is then given by
\begin{align*}
    \ell(\theta,Z_i) := -(Y_i\log (\sigma(\theta^TX_i)) + (1-Y_i)\log (1 - \sigma(\theta^TX_i))),
\end{align*}
where $\sigma(x) = \frac{1}{1+e^{-x}}$. Let $\theta^*_t = (0.3,0.5)$ and $\theta^*_s = (0.2,0.4)$ denote the true parameters for the target and source domains. Given $m = 5000$, let the marginal prior $\omega(\Theta_s)$ be uniformly distributed over $[0,1]^2$ and our prior knowledge $\omega(\Theta_t|\Theta_s)$ assumes that $\Theta_t$ is normally distributed with the mean of $\Theta_s$ and covariance of $\begin{bmatrix}
c^2 & 0\\
0 & c^2
\end{bmatrix}$, here $c$ represents the prior belief on $\Theta_t$ such that smaller $c$ implies $\Theta_t$ is closer to $\Theta_s$ and vice versa. To show the usefulness of the source data, we compare with the target only case ($m = 0$) where we assume the prior $\hat{\omega}(\Theta_t)$ is uniformly distributed over $[0,1]^2$. 

\begin{figure}[h!]
\centering
\subfigure{\includegraphics[width = 1.12in]{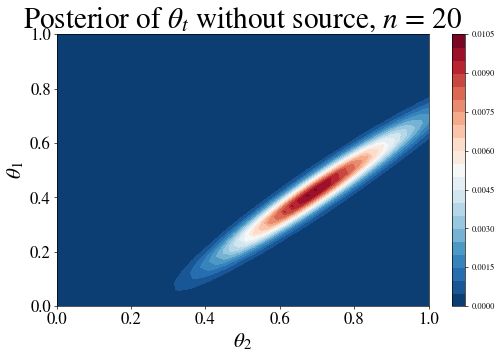}}
\subfigure{\includegraphics[width = 1.12in]{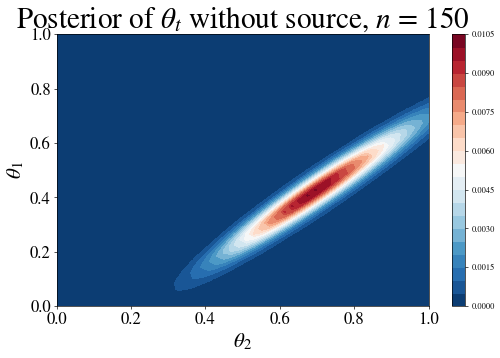}}
\subfigure{\includegraphics[width = 1.12in]{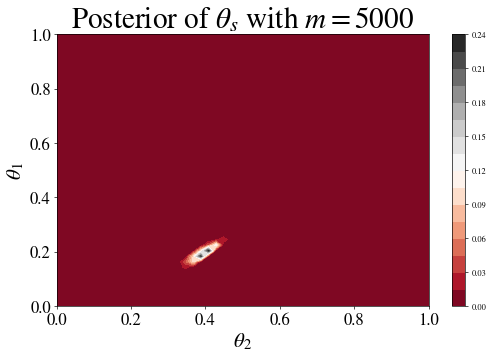}} \\
\subfigure{\includegraphics[width = 1.12in]{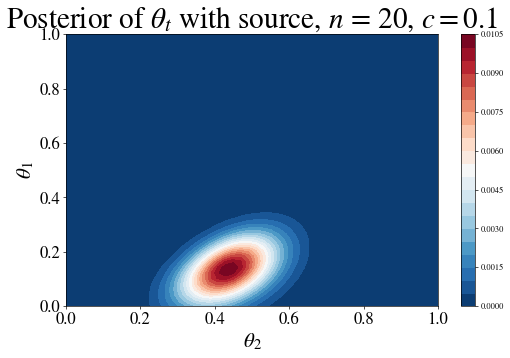}} 
\subfigure{\includegraphics[width = 1.12in]{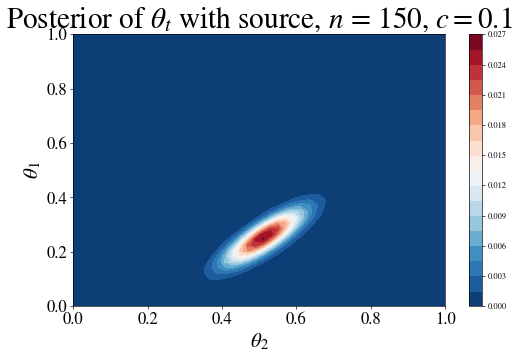}}
\subfigure{\includegraphics[width = 1.12in]{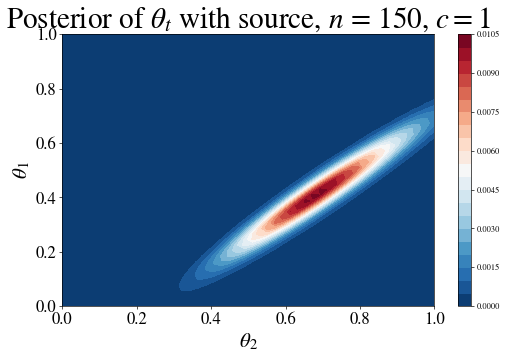}}
\caption{The posterior of $\theta_s$ and $\theta_t$ given $D^m_s$ and $D^n_t$ under different prior belief $c$ and target sample size $n$}
\label{fig:post}
\end{figure}

After receiving $n$ target samples, we plot different posteriors to see the effect of the mixture strategy induced by the chosen prior. From Figure~\ref{fig:post}, given sufficient source data, the posterior of $\Theta_s$ will give a precise estimation of $\theta^*_s$ and the density will mostly concentrate around $[0.2,0.4]$. While there is a lack of target samples ($n$ is  small), the posterior $\hat{Q}(\Theta_t|D^n_t)$ without the source is relatively scattered and the density around $\theta^*_t$ is quite low. On the contrary, with the prior knowledge $\omega(\Theta_t|\Theta_s)$ and small $c = 0.1$, the posterior $Q(\Theta_t|D^m_s, D^n_t)$ will be concentrated more around $\theta^*_t$ as source and target parameters are particularly close. When $c$ increases to $1$, the source data is no longer helpful as $\Theta_t$ is roughly distributed uniformly on $[0,1]^2$ and the posterior behaves similarly to target only case.

\begin{figure}[h!]
    \centering
    \subfigure[Positive Transfer]{\includegraphics[width = 1.72in]{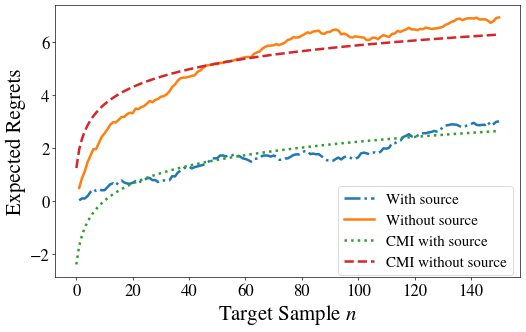}}
    \subfigure[Negative Transfer]{\includegraphics[width = 1.72in]{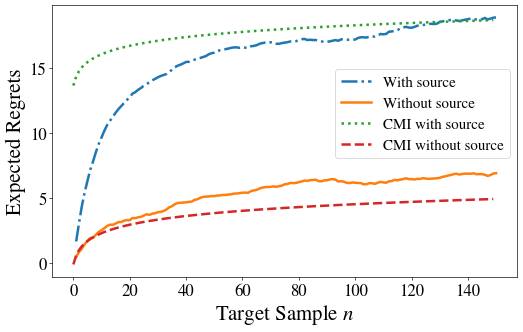}}
    \caption{The comparisons of the expected regret $\mathcal{R}(n)$ of the positive transfer with $\theta^*_s = [0.2,0.4]$ (left) and the negative transfer (right) with $\theta^*_s = [0.8,0.2]$ under the common settings where $\theta^*_t = [0.3,0.5]$ and $c = 0.1$. The results are averaged over 200 experiments.}
    \label{fig:pos_transfer}
\end{figure}

 To further demonstrate our theoretical results, we plot the expected regrets in Figure~\ref{fig:pos_transfer} for positive and negative transfer cases, and we also plot the asymptotic estimation of CMI in dashed lines from Theorem~\ref{thm:cmi} and \ref{theorem:gene-para} to numerically evaluate the difference. From the left figure, it is observed that introducing the source indeed yields lower regret, which fits our intuition from the posteriors. Even for small $n$($\approx 40$), CMI captures the regret quite well and the gap is roughly $\log\frac{\omega(\theta^*_t|\theta^*_s)}{\omega(\theta^*_t)}$ as noted in Remark~\ref{remark:consist}. In contrast, we also examine the negative transfer case with $\theta^*_s = [0.8,0.2]$ where the results are shown in the right figure. With this specific choice of $\theta^*_s$, the prior distribution $\omega(\theta^*_c|\theta^*_s)$ in this case has an extremely low density and the estimation will hardly approach the true parameters. As a result, the negative transfer happens and source samples will hurt the performance instead. It also appears that CMI captures this trend well when $n$ goes reasonably large ($\approx 80$). 

Overall, in both positive and negative transfer cases, the  gaps between the regrets are mainly reflected on the prior knowledge $\omega(\theta^*_t|\theta^*_s)$ when $n$ is reasonably large as mentioned in Remark~\ref{remark:consist} and~\ref{remark:knowledge_transfer}, which experimentally confirms Theorem~\ref{theorem:gene-para}. Moreover, it shows that the asymptotic bounds are still reasonably accurate in the case when $n$ and $m$ are small. 
\newpage
\bibliographystyle{IEEEtran}
\bibliography{reference}

\IEEEtriggeratref{3}

\end{document}